\documentclass{bmvc2k}

\usepackage{times}
\usepackage{epsfig}
\usepackage{graphicx}
\usepackage{amsmath}
\usepackage{amssymb}
\usepackage{sidecap}
\usepackage{enumitem}
\usepackage{hyperref} 

\title{Bipartite Conditional Random Fields for Panoptic Segmentation}

\addauthor{Sadeep Jayasumana}{sadeep@apache.org}{1}
\addauthor{Kanchana Ranasinghe}{kahnchana@gmail.com}{2}
\addauthor{Sahan Liyanarachchi}{150360a@uom.lk}{2}
\addauthor{Mayuka Jayawardhana}{150273j@uom.lk}{2}
\addauthor{Harsha Ranasinghe}{150504v@uom.lk}{2}
\addauthor{Sina Samangooei}{sina@five.ai}{3}

\addinstitution{
	Work done at Five AI
}

\addinstitution{
  University of Moratuwa, \\
  Moratuwa, Sri Lanka
}

\addinstitution{
	Five AI, \\
	Cambridge, UK
}

\runninghead{Jayasumana et al}{BCRF for Panoptic Segmentation}

\renewcommand{\vec}[1]{\mathbf{#1}}

\usepackage{caption}
\usepackage{tabularx}
\usepackage{multirow}
\usepackage{amsmath}
\usepackage{amssymb}
\usepackage{graphicx}
\usepackage{algorithm}% http://ctan.org/pkg/algorithms
\usepackage{algpseudocode}

\usepackage{amsthm}

\renewcommand{\vec}[1]{\mathbf{#1}}

\newcommand{\Pro}{\operatorname{Pr_0}}
\newcommand{\X}{\mathbf{X}}

\newcommand{\Z}{\mathbf{Z}}

\newcommand{\inst}{\texttt{inst}}
\newcommand{\x}{\vec{x}}

\newcommand{\z}{\vec{z}}
\newcommand{\Lstuff}{\mathcal{L}_{\operatorname{stuff}}}
\newcommand{\Lthings}{\mathcal{L}_{\operatorname{things}}}

\newcommand{\Sim}{\operatorname{Sim}}

\newcommand{\minuseq}{\mathrel{{-}{=}}}

\newcommand{\coleq}{\mathrel{{:}{=}}}
\newcommand{\cl}{\operatorname{class}}

\begin{document}

\maketitle

%%%%%%%%% ABSTRACT
\vspace{-0.4cm}
\begin{abstract}
   We tackle the panoptic segmentation problem with a conditional random field (CRF) model. Panoptic segmentation involves assigning a semantic and an instance label to each pixel of a given image. At each pixel, the semantic label and the instance label should be compatible. Furthermore, a good panoptic segmentation should have a number of other desirable properties such as the spatial and color consistency of the labeling. To tackle this problem, we propose a CRF model, named Bipartite CRF or BCRF, with two types of random variables for semantic and instance labels. In this formulation, various energies are defined within and across the two types of random variables to encourage a consistent panoptic segmentation. We propose a mean-field-based efficient inference algorithm for solving the CRF and empirically show its convergence properties. This algorithm is fully differentiable, and therefore, BCRF inference can be included as a trainable module in any deep network. In the experimental evaluation, we quantitatively and qualitatively show that the BCRF yields superior panoptic segmentation results in practice. Our code is publicly available at: \href{https://github.com/sahan-liyanaarachchi/bcrf-detectron}{https://github.com/sahan-liyanaarachchi/bcrf-detectron}. 
\end{abstract}
\vspace{-0.4cm}
%%%%%%%%% BODY TEXT
\section{Introduction}
\vspace{-0.2cm}
Panoptic segmentation of images is a problem that has received considerable attention in computer vision recently. It combines two well-known computer vision tasks: semantic segmentation and instance segmentation. The goal of panoptic segmentation is to assign a semantic label and an instance label for each pixel in the image as presented formally in \cite{panoptickirillov2017}.

 Although semantic segmentation and instance segmentation are apparently very related problems, current state of the art methods in computer vision solve these in substantially different ways. The semantic segmentation problem is usually solved with a fully convolutional network architecture such as FCN~\cite{fcn_pami2017} or DeepLab~\cite{Deeplab_pami}, whereas the instance segmentation problem is solved using an object detector based method such as Mask-RCNN~\cite{mask_rcnn}. Each of these architectures have their own strengths and weaknesses. For example, fully-convolutional network based semantic segmentation methods have a wide field of view, specially when used with dilated convolutions~\cite{dialated_conv}, and therefore can make semantic segmentation predictions with global information about the image. In contrast, region proposal based networks, such as Mask-RCNN, focus on specific regions of interest during the later stages of the network and make predictions using strong local features available within a given region of interest \cite{bagnets}. It is natural to think of a systematic way of combining the complementary strengths of these two different approaches.

We propose a Conditional Random Field (CRF) based framework for panoptic segmentation: Bipartite Conditional Random Fields (BCRF). This module performs probabilistic inference on a graphical model to obtain the best panoptic label assignment given the semantic segmentation classifier, the instance segmentation classifier, and the image itself. Our framework provides a heuristic-free, probabilistic method to combine semantic segmentation results and instance segmentation results - yielding a panoptic segmentation with consistent labeling across the entire image. We formulate our bipartite CRF using different energy functions to encourage the spatial, appearance and semantic consistency of the final panoptic segmentation. The optimal labeling is then obtained by performing mean field inference on the bipartite CRF - solving for both the semantic segmentation and the instance segmentation in a jointly optimal way.

Importantly, we show that our proposed BCRF inference is fully differentiable with respect to the parameters used within the CRF and also the semantic segmentation and instance segmentation classifier inputs. Therefore, the BCRF module can be used as a first-class citizen of a deep neural network to perform panoptic segmentation. A deep network equipped with the BCRF module is capable of structured prediction of consistent panoptic labels and is end-to-end trainable. We show an example application of this framework and demonstrate that superior results can be gained by probabilistic combination of a semantic segmentation classifier and an instance segmentation classifier in the BCRF framework.

\begin{figure*}[t!]
\vspace{0.3cm}
  \centering
  \includegraphics[trim = 1mm 1mm 1mm 1mm, clip, width=\textwidth]{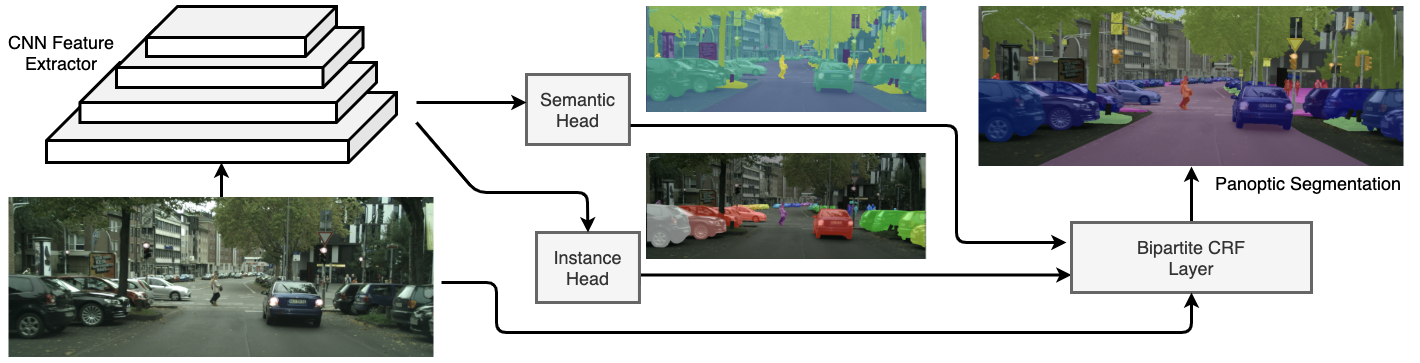}
  \vspace{0.1cm}
  \caption{{\bf BCRF in an end-to-end trainable deep net.} \small{The Bipartite CRF proposed in this paper can be used to combine the predictions of a semantic segmentation model and an instance segmentation model to obtain a consistent panoptic segmentation.}}
  \label{fig:bcrf_net}
\vspace{-0.4cm}
\end{figure*}

\vspace{-0.2cm}
\section{Related Work}
\vspace{-0.2cm}
\subsection{Panoptic Segmentation}
\vspace{-0.1cm}
Since its formal introduction by Kirillov \emph{et al.}~\cite{panoptickirillov2017}, the task of panoptic segmentation has gained popularity, with multiple works attempting to transform existing network architectures to tackle this task~\cite{panoptic_spatial_ranking, panoptic_DeeperLab, panoptic_SSAP, panoptic_attention, panoptic_scene, Upsnet_paper}. The work in \cite{Upsnet_paper} presents a parameter-free panoptic head that logically combines instance and semantic logits. Our work achieves a different goal of individually optimising the two sets of logits learning arbitrary complex mappings between them. Also, the panoptic combination head in \cite{Upsnet_paper} could be used on top of our module instead of the generic combination in \cite{panoptickirillov2017} that we use for further improvement. The spatial ranking methods described in \cite{panoptic_spatial_ranking, InstanceOcclusion} optimize ranking between overlapping instance masks. The work in \cite{panoptic_attention} uses two attention modules to optimize the background segmentation. Our framework performs both these tasks together using our cross-potential terms while enforcing the two branches to have a consensus in their outputs. The BCRF module is thus more robust in terms of information integration.  

Another similar recent work by Arnab \emph{et al.}~\cite{Anurag17} moves in a slightly new direction by using a CRF to obtain instance segmentation outputs from a semantic segmentation using bounding box (from an object detection network) and instance shape cues. Our work differs from this in three significant ways: presence of pixel-wise cross potentials, using instance mask cues from a region-based network, and the ability to explicitly learn and model relationships between classes. 

\vspace{-0.3cm}
\subsection{Conditional Random Fields}
\vspace{-0.1cm}
Conditional Random Fields (CRFs) are a class of statistical modeling models excellent at structured prediction tasks such as semantic segmentation. While early methods of CRFs for semantic segmentation ~\cite{instanceCRF01, instanceCRF02} used 4-connected or 8-connected locally connected graphs, the development of an efficient mean field based inference algorithm ~\cite{densecrf} to solve fully connected CRFs with Gaussian edge potentials resulted in a resurgence of its use in deep networks. The authors of~\cite{Zhen_ICCV15_CRFRNN} showed that this CRF inference algorithm can be formulated as a Recurrent Neural Network (RNN), which plugged into a fully convolutional network could obtain the state-of-the-art in semantic image segmentation. Similar trainable CRF models have been used in works such as \cite{arnab_eccv_2016, convCRF} for semantic segmentation and ~\cite{Anurag17} for instance segmentation. In~\cite{li_eccv_2018}, where the problem of panoptic segmentation with weak and semi supervision was addressed, the authors used a CRF for refining instance segmentation labels. However, it worked on homogeneous instance labels only and therefore was similar in spirit to previous fully connected CRFs.

In our work, we propose a bipartite CRF operating on the semantic segmentation task and the instance segmentation task \emph{simultaneously}. This CRF has energies within semantic segmentation labels, energies within instance segmentation labels, and also energies \emph{across} semantic and instance segmentation labels. To the best of our knowledge, this is the first time a bipartite CRF with cross connections between semantic and instance labels has been proposed in the context of pixel-wise labeling. 
\begin{table*}[hbt!]
\vspace{-0.0cm}
%  \centering
	\begin{subtable}
        \centering
        {\renewcommand{\arraystretch}{0.7}
                                \begin{tabular}{ c@{\hspace{3pt}} c@{\hspace{3pt}} c@{\hspace{3pt}} c@{\hspace{3pt}} c}				
						\includegraphics[width=0.19\textwidth]{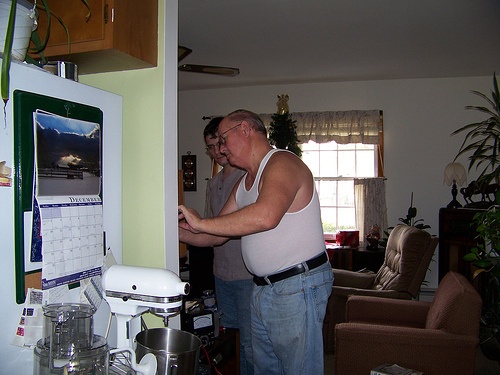}%
						& \includegraphics[width=0.19\textwidth]{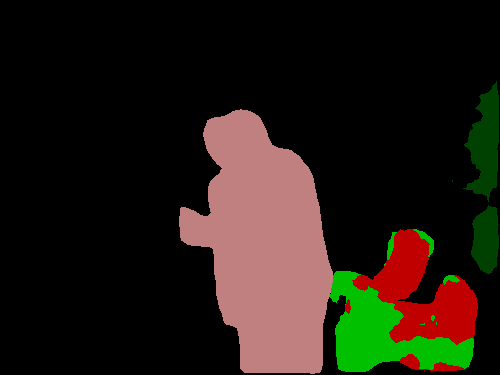}%
						& \includegraphics[width=0.19\textwidth]{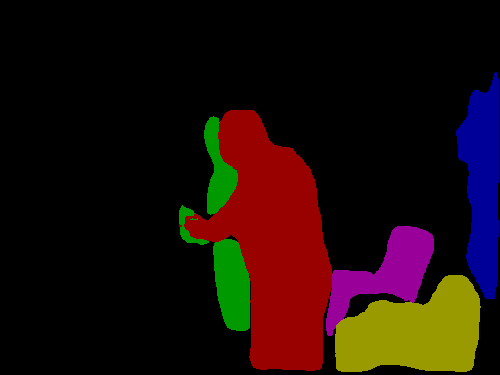}%
						& \includegraphics[width=0.19\textwidth]{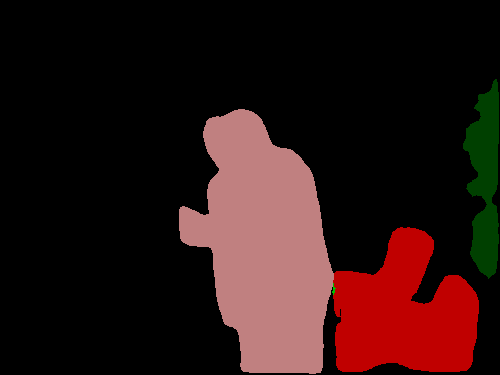}%
						& \includegraphics[width=0.19\textwidth]{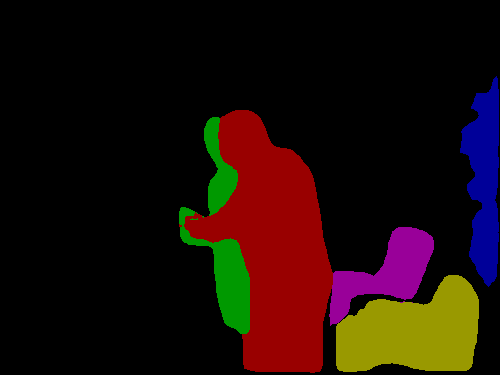}\\
														\newline
%						\includegraphics[width=0.19\textwidth]{figs/pascal/original/421.jpg}%
%						& \includegraphics[width=0.19\textwidth]{figs/pascal/before_sem/421.png}%
%						& \includegraphics[width=0.19\textwidth]{figs/pascal/before_ins/421.png}%
%						& \includegraphics[width=0.19\textwidth]{figs/pascal/after_sem/421.png}%
%						& \includegraphics[width=0.19\textwidth]{figs/pascal/after_ins/421.png}\\
%						                                \newline
						\includegraphics[width=0.19\textwidth]{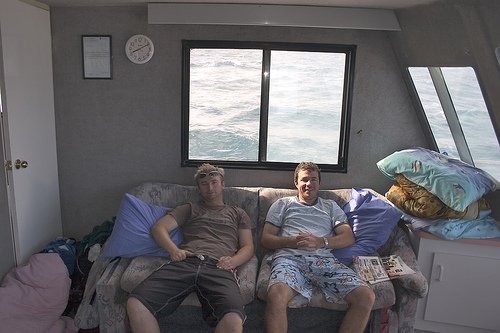}%
						& \includegraphics[width=0.19\textwidth]{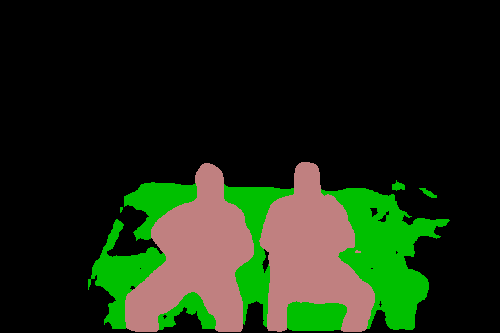}%
						& \includegraphics[width=0.19\textwidth]{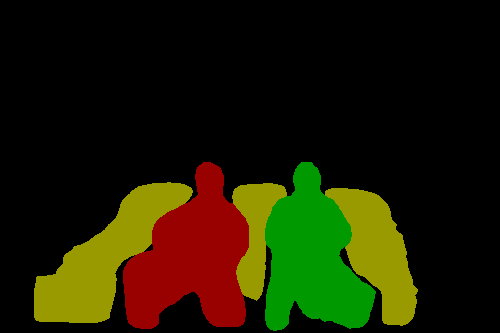}%
						& \includegraphics[width=0.19\textwidth]{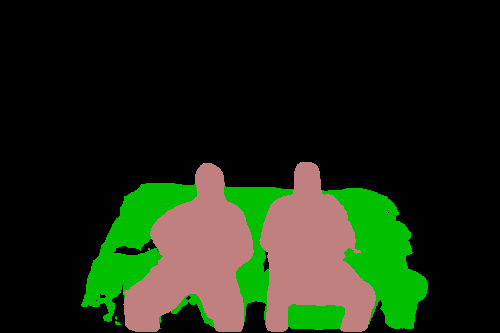}%
						& \includegraphics[width=0.19\textwidth]{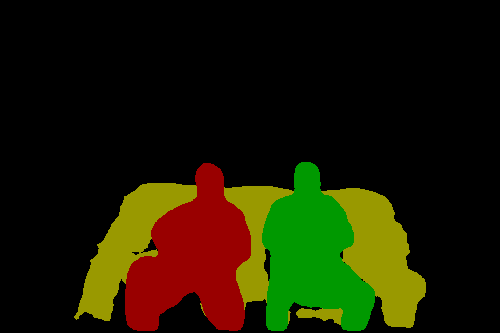}\\
						                               \newline
%						\includegraphics[width=0.19\textwidth]{figs/pascal/original/468.jpg}%
%						& \includegraphics[width=0.19\textwidth]{figs/pascal/before_sem/468.png}%
%						& \includegraphics[width=0.19\textwidth]{figs/pascal/before_ins/468.png}%
%						& \includegraphics[width=0.19\textwidth]{figs/pascal/after_sem/468.png}%
%						& \includegraphics[width=0.19\textwidth]{figs/pascal/after_ins/468.png}\\
%						\newline
%						\includegraphics[width=0.19\textwidth]{figs/pascal/original/184.jpg}%
%						& \includegraphics[width=0.19\textwidth]{figs/pascal/before_sem/184.png}%
%						& \includegraphics[width=0.19\textwidth]{figs/pascal/before_ins/184.png}%
%						& \includegraphics[width=0.19\textwidth]{figs/pascal/after_sem/184.png}%
%						& \includegraphics[width=0.19\textwidth]{figs/pascal/after_ins/184.png}\\
%						\newline
%						\includegraphics[width=0.19\textwidth]{figs/pascal/original/226.jpg}%
%						& \includegraphics[width=0.19\textwidth]{figs/pascal/before_sem/226.png}%
%						& \includegraphics[width=0.19\textwidth]{figs/pascal/before_ins/226.png}%
%						& \includegraphics[width=0.19\textwidth]{figs/pascal/after_sem/226.png}%
%						& \includegraphics[width=0.19\textwidth]{figs/pascal/after_ins/226.png}\\
%						                                 \newline
 						\includegraphics[width=0.19\textwidth]{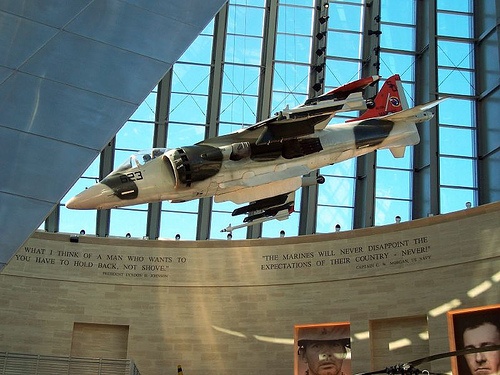}%
                          & \includegraphics[width=0.19\textwidth]{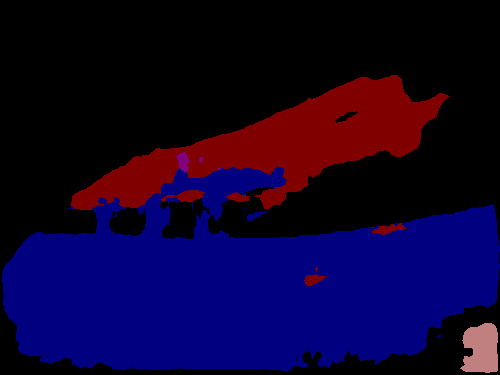}%
                          & \includegraphics[width=0.19\textwidth]{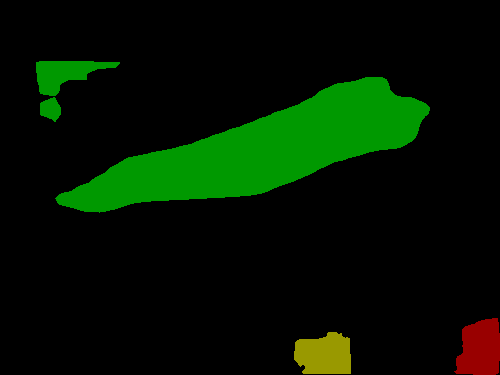}%
                          & \includegraphics[width=0.19\textwidth]{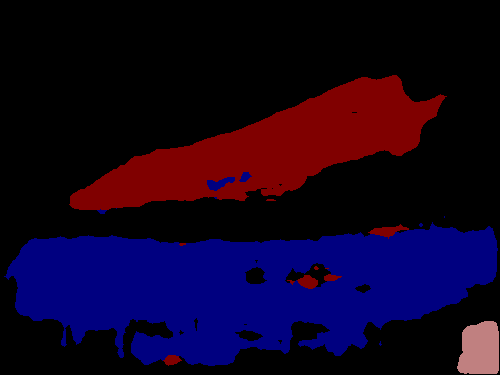}%
                          &\includegraphics[width=0.19\textwidth]{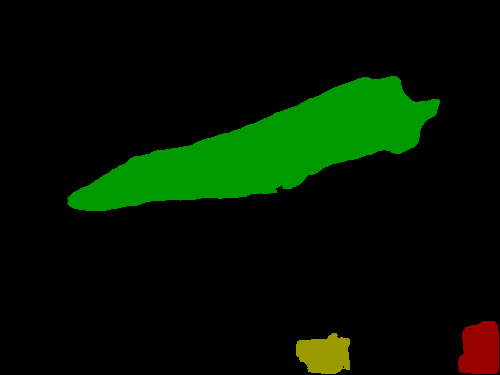}\\			
                                \end{tabular}
                                }
    \end{subtable}
    \vspace{0.2cm}
    \caption{\small{{\bf Visualizations on Pascal VOC Dataset.} Columns left to right: original image, semantic output and instance output before BCRF, semantic output and instance output after BCRF.}}
    \label{tbl:pascal_visual}
    \vspace{-0.3cm}
\end{table*}

\section{Background: Conditional Random Fields}
\vspace{-0.0cm}
A CRF, used in the context of pixel-wise label prediction, models pixel labels as random variables that form a Markov Random Field (MRF) when conditioned upon the image. CRFs have primarily been used in computer vision for semantic image segmentation. In this setting, CRFs encourage the desirable properties of a good segmentation, such as the spatial consistency (e.g. spatially neighboring pixels should have the same label) and color consistency (e.g. a semantic segmentation boundary should correspond to an edge in the image) through various energy functions used in the formulation. A CRF formulation usually has energy terms arising from an imperfect classifier (sometimes known as the unary energy) and energy terms encouraging the consistency properties of the segmentation (sometimes known as the pairwise energy). Some semantic CRF models also include higher order energy terms to encourage higher order consistency properties such as consistency of the labeling within super-pixels~\cite{arnab_eccv_2016}.

Once an appropriate energy function is formed, the optimal labeling is found as the labeling that minimizes the CRF energy (or equivalently, maximizes the probability). This is known as the inference of the CRF. The exact inference of a CRF with dense pairwise connections is intractable and hence approximate inference methods such as mean field variational inference has to be utilized to solve the CRF in reasonable time~\cite{densecrf}. For a detailed treatment of CRFs, the reader is referred to~\cite{Koller_book}.  
\vspace{-0.2cm}
\section{Bipartite CRFs}
\label{sec:body}
\vspace{-0.1cm}
We propose a CRF formulation with bipartite random variables to capture interactions between semantic labels and instance labels. Inference of this CRF gives the jointly most probable semantic and instance segmentation (and therefore, the panoptic segmentation) for a given image. 

For each pixel $i$, define a pair of discrete random variables $(X_i, Z_i)$ to denote
its semantic label and the instance label, respectively. For each $i$, $X_i$ can take values in $\mathcal{L} = \{l_1, l_2, \dots, l_L\}$, where each $l_j$ is a semantic label and $L$ is the number of semantic labels (includes both stuff and thing classes). Therefore, $\mathcal{L} = \Lstuff \cup \Lthings$, where $\Lstuff$ is the set of stuff class labels and $\Lthings$ the set of thing class labels. Similarly, for each $i$, $Z_i$ can take values in $\mathcal{T} = \{\inst_0, \inst_1, \dots, \inst_{N_{inst}} \}$, where $N_{inst}$ is the number of instances detected in the image, and the label $\inst_0$ is reserved to represent the "no instance'' case (the pixel belongs to a stuff class).

Let $\X = [X_1, X_2, \dots, X_N ]$ and $\Z = [Z_1, Z_2, \dots, Z_N
]$, where $N$ is the number of the pixels in the image. A joint assignment $(\x, \z)$ to these two random vectors $(\X, \Z)$ gives a unique semantic label and an instance label to each pixel $i$, and therefore represents a panoptic segmentation of the image. Note that, $\x \in \mathcal{L}^N$ and $\z \in \mathcal{T}^N$. In this work, we discuss the probability of such assignments and formulate the probability distribution function so that the ``good'' panoptic segmentation will have a high probability. We then perform inference on this formulation to find the assignment that maximizes the probability to obtain the best panoptic segmentation.

The probability of a panoptic segmentation $(\x, \z)$, given the image $I$, can be modeled as a Gibbs distribution of the following form:
\vspace{-0.3cm}
\begin{equation}
\label{eqn:prob}
\Pr(\X = \x, \Z = \z|I) = \frac{1}{\mathcal{Z}(I)}\exp(-E(\x, \z|I)),
\vspace{-0.2cm}
\end{equation}
where $\mathcal{Z}(I) = \sum_{(\x,\z)} \exp(-E(\x, \z|I))$, is a normalization constant, sometimes known as the partition function. The term $E(\x, \z|I)$ is known as the energy of the configuration $(\x, \z)$. Hereafter, we drop the conditioning on $I$ in the notation for brevity. The energy of our bipartite CRF is defined as follows:
\vspace{-0.1cm}
\begin{equation}
\label{eqn:energy}
\begin{split}
E(\x, \z) =& \sum_i\phi(x_i) + \sum_{i < j}\Phi(x_i, x_j) \;+  \sum_i\psi(z_i) + \sum_{i < j}\Psi(z_i, z_j) \; + \sum_i\omega(x_i, z_i) + \sum_{i < j}\Omega(x_i, z_j)
\end{split}
\end{equation}

\noindent where $x_i$ and $z_i$ are the elements of the vectors $\x$ and $\z$, respectively. The meaning of each term will be described in detail below. Note that, since a ``good'' panoptic segmentation should have a high probability, it should have a low energy. Various terms in Eq.~\eqref{eqn:energy} should therefore encourage a good panoptic segmentation by penalizing disagreements with our prior knowledge about a consistent panoptic segmentation. 

\subsection{Semantic \& Instance Components of the CRF}
In the following, we discuss the first two terms of the energy function in Eq.~\eqref{eqn:energy}. The first term encourages the semantic segmentation result to be consistent with the initial classifier,
\vspace{-0.2cm}
\begin{equation}
\label{eqn:detector}
\phi(X_i = x_i) = - \log(\Pro(X_i = x_i)),
\vspace{-0.2cm}
\end{equation}
\noindent where $\Pro(.)$ is the classifier probability score for the semantic segmentation. The second term in Eq.~\eqref{eqn:energy} encourages the smoothness of the semantic labeling,
\vspace{-0.2cm}
\begin{equation}
\label{eq:similarity}
\Phi(X_i = x_i, X_j = x_j) = \mu(x_i, x_j) \operatorname{Sim_\Phi}(i, j),
\end{equation}
where $\mu: \mathcal{L} \times \mathcal{L} \to \mathbb{R}$ is the label compatibility function, and $\operatorname{Sim_\Phi}(i, j)$ is a similarity measure between the pixels $i$ and $j$. This term penalizes assigning different labels to a pair of pixels that are "similar". Following \cite{densecrf}, we use a mixture of Gaussians as the similarity measure and define a general similarity function,
\vspace{-0.2cm}
\begin{equation}
\label{eqn:kernels}
\operatorname{Sim_\chi}(i, j) = \sum_m w_{\chi, m} \exp\left(-\frac{\|\vec{f}_i^{(m)} - \vec{f}_j^{(m)}\|^2}{2\sigma_{\chi,m}^2}\right)
\vspace{-0.2cm}
\end{equation}
where $\mathbf{f}_i$ is a feature vector for pixel $i$ containing information such as its spatial location and bilateral features (RGB + spatial coordinates). We use the same spatial and bilateral features used in~\cite{densecrf}. The similarity measure $Sim_\Phi$ is derived accordingly. 

The next two terms of the energy function in Eq.~\eqref{eqn:energy} perform the same for instance classification. Similarly we assume the existence of an initial classifier, such as Mask R-CNN. Despite methods like Mask R-CNN providing fixed-size predictions with respect to the bounding boxes of the detections, these predictions can be easily mapped to the full image by using bilinear interpolation and trivial coordinate transforms similar to \cite{Upsnet_paper} as follows. 

If there are N detections in the MaskRCNN output, instance segmentation unary potentials is a tensor of shape $[im_{height}, im_{width}, N + 1]$. There are (N + 1) channels to accommodate no instance at channel 0, i.e. at each pixel, the unary potential is a vector of length (N + 1) that contains negative logits (see Eq.~\eqref{eqn:detector}) corresponding to the detection confidence of each particular instance. Each pixel in the image may belong to none, one or multiple instances, since overlaps between bounding boxes are possible. These cases are handled as follows: 1) None: "no-instance" channel (channel 0) will have a high negative score. 2) One: The corresponding channel will have a negative score proportional to the detection confidence of the MaskRCNN. 3) Multiple (say $u$ and $v$): channels $u$ and $v$ will have negative scores proportional to the confidence scores for the two detections. In all cases, other channels will be set to zero.

Similar to the semantic segmentation case, the third term in Eq.~\eqref{eqn:energy} encourages the panoptic segmentation to be consistent with the instance classifier probabilities. The fourth term in Eq.~\eqref{eqn:energy} encourages instance label consistency across the whole image by penalizing assigning different instance labels to similar pixels:
\vspace{-0.2cm}
\begin{equation}
\Psi(Z_i = z_i, Z_j = z_j) = [z_i \neq z_j] \operatorname{Sim_\Psi}(i, j).
\vspace{-0.2cm}
\end{equation}
The compatibility transform in this case is fixed to be $[z_i \neq z_j]$, where $[.]$ is the Iverson bracket. The similarity measure $\Sim_\Psi$ is derived from Eq.~\eqref{eqn:kernels}.

\subsection{Cross Potentials in the CRF}
An important contribution of this paper is the introduction of cross potentials between the semantic segmentation and instance segmentation. The semantic segmentation and the instance segmentation are highly related problems and therefore the solutions should agree: the semantic label at any pixel has to be compatible with the instance label at that pixel. For example, if the instance labeling says that the pixel $i$ belongs to an instance of a person class, the semantic label at pixel $i$ should also have the person label. If the initial classifier results for the instance segmentation and the semantic segmentation do not agree, one of them should correct itself depending on the interactions of other terms in the CRF.

The first cross potential term (the fifth term in Eq.~\eqref{eqn:energy}), encourages instance label and the semantic label at a given pixel to agree:
\vspace{-0.1cm}
\begin{equation}
\omega(X_i = x_i, Z_i = z_i) = f(x_i, \cl(z_i)).
\vspace{-0.1cm}
\end{equation}
Here, $\cl(z_i)$ is the class label of the instance $z_i$ with $\inst_0$ mapped to a special class $\operatorname{null}$. Note that, for all valid instances, the class label can be obtained from the instance classifier (e.g. Mask R-CNN). The function $f(., .): (\mathcal{L}, \Lthings \cup \{\operatorname{{null}}\}) \to \mathbb{R}^+_0$, captures the cost of incompatibility and is defined as follows: 
\vspace{-0.2cm}
\begin{equation}
\label{eqn:cross_compat}
f(x_i, \cl(z_i)) = \begin{cases}
0,\;\;\text{if}\;x_i = \cl(z_i)\\
0,\;\;\text{if}\;x_i \in \mathcal{L}_{\text{stuff}}\;\text{and}\;\cl(z_i) = \operatorname{null}\\
\eta(x_i, \cl(z_i)),\;\; \text{otherwise}.
\end{cases}
\vspace{-0.1cm}
\end{equation}
The above function covers three cases: 1) If the semantic label and the class label of the instance label match, there will be no penalty for such assignment since there is no incompatibility in this case. 2) If the semantic segmentation assigns a stuff label and the instance segmentation assigns $\inst_0$ label, there will be no penalty in that case either. 3) If the semantic label and the instance label mismatch, there will be a penalty with the magnitude decided by the function $\eta(., .): \Lthings \cup \{\operatorname{null}\} \times \Lthings \cup \{\operatorname{null}\} \to \mathbb{R}^+$. This function is learned from data as described in Section~\ref{sec:infer}.

The last term in Eq.~\eqref{eqn:energy}, encourages the consistency of semantic label and the instance label among similar looking pixels and has the form:
\vspace{-0.2cm}
\begin{equation}
\Omega(X_i = x_i, Z_j = z_j) = f(x_i, \cl(z_j))\; \operatorname{Sim_\Omega}(i, j),
	\vspace{-0.1cm}
\end{equation}
where each symbol has the meaning described above and $Sim_\Omega$ is derived from Eq.~\eqref{eqn:kernels}.
\section{Inference and Parameter Optimization}
\label{sec:infer}
\vspace{-0.1cm}
The best panoptic segmentation given the model described in Section~\ref{sec:body} is the assignment $(\x, \z)$ that maximizes the probability in Eq.~\eqref{eqn:prob}. However, since the graphical model used in BCRF has dense connections between the pixels, the exact inference is infeasible. We therefore use an approximate parallel mean field inference algorithm following~\cite{densecrf}.

In this setting, the joint probability distribution is approximated by the product of marginal distributions:
\vspace{-0.2cm}
\begin{equation}
\label{eq:q_approx}
\Pr(\X=\x, \Z=\z) \approx \prod_i Q_i(x_i)\,R_i(z_i),
\vspace{-0.2cm}
\end{equation}
where $Q_i(x_i) = \Pr(X_i = x_i)$ and $R_i(z_i) = \Pr(Z_i = z_i)$ are the marginal distributions. Out of all the distributions that can be written down in this factorized form, the closest distribution to the original joint distribution is found by minimizing the KL divergence~\cite{Koller_book, densecrf}. For our BCRF formulation, this results in the iterative algorithm detailed in Algorithm~\ref{alg:infer}.

\vspace{-0.2cm}
\begin{algorithm*}
  \caption{Inference on Bipartite CRF \label{alg:infer}}
  \begin{algorithmic}[1]
  	  %\Statex
      \State $Q_i(l) \coleq \operatorname{softmax}_i(-\phi_i(l))$ and $R_i(t) \coleq \operatorname{softmax}_i(-\psi_i(t))$\Comment{Initialization}
      \While{not converged}
%        \State $\tilde{Q}_i(l) \gets \sum_{j \neq i}{\Sim_\Phi(i, j)\,Q_j(l)}$ 
%        \State $\tilde{R}_i(t) \gets \sum_{j \neq i}{\Sim_\Psi(i, j)\,R_j(t)}$
%        \State $\hat{Q}_i(l) \gets -\sum_{l' \in \mathcal{L}} \mu(l, l')\;\tilde{Q}_i(l')$
%        \State $\hat{R}_i(t) \gets -\sum_{t' \in \mathcal{T}} [t \neq t']\;\tilde{R}_i(t')$
        \State $Q'_i(l) \minuseq \phi_i(l)$ \Comment{Update due to the first term}
        \State $Q'_i(l) \minuseq \sum_{l' \in \mathcal{L}} \left(\mu(l, l')\sum_{j \neq i}{\Sim_\Phi(i, j)\,Q_j(l')}\right)$ \Comment{Update due to the second term}
        
        \State $R'_i(t) \minuseq \psi_i(t)$ \Comment{Update due to the third term}
        \State $R'_i(t) \minuseq \sum_{t' \in \mathcal{T}} \left([t \neq t']\sum_{j \neq i}{\Sim_\Psi(i, j)\,R_j(t')}\right)$ \Comment{Update due to the fourth term}
        \State $Q'_i(l) \minuseq \sum_{t \in \mathcal{T}} \Big( f(l, \cl(t))\, R_i(t)\Big)$
        \State $R'_i(t) \minuseq \sum_{l \in \mathcal{L}} \Big( f(l, \cl(t))\, Q_i(l) \Big)$ \Comment{Updates due to the fifth term}

		\State $Q'_i(l) \minuseq \sum_{t \in \mathcal{T}} \left( f(l, \cl(t))\, \sum_{j \neq i}{\Sim_\Omega(i, j)\,R_j(t')} \right)$
		\State $R'_i(t) \minuseq \sum_{l \in \mathcal{L}} \left( f(l, \cl(t))\, \sum_{j \neq i}\Sim_\Psi(i, j)\,Q_j(l')\right)$ \Comment{Updates due to the sixth term}
		\State $Q_i(l) \coleq \operatorname{softmax}_i\Big(Q'_i(l)\Big)$ and $R_i(t) \coleq \operatorname{softmax}_i\Big(R'_i(t)\Big)$ \Comment{Normalization}
      \EndWhile
      %\State
  \end{algorithmic}
\end{algorithm*}

\vspace{-0.2cm}

To make our model flexible, we deliberately include a number of parameters in the BCRF model, which we automatically learn from the training data. More specifically, the BCRF model has the following parameters:

\begin{enumerate}[topsep=0pt,itemsep=-0ex,partopsep=1ex,parsep=0ex]

\item Weight multipliers for different energy terms: each term in Eq.~\eqref{eqn:energy} is multiplied with a weight parameter, which decides the relative strength of the term. This parameterization helps learn the optimal combination of different energies in the CRF. For example, if the initial semantic segmentation model has better accuracy than the instance segmentation model, the $\phi$ unary energy might be weighted more than the $\psi$ unary energy.

\item Parameters for similarity functions: Each similarity function $\Sim_X(i, j)$ of the form shown in Eq.~\eqref{eqn:kernels} has its own parameters. These learn the relative strength of spatial and appearance consistency of the panoptic segmentation.

\item Label compatibility matrices: The two functions $\mu(., .)$ and $\eta(., .)$ are initialized to have a zero cost for a pair of identical labels and a fixed cost for any combination of two different labels. They are then given the freedom to automatically learn the relative penalty strengths for different label combinations.
\end{enumerate}
\vspace{-0.3cm}
\vspace{-0.2cm}
\section{Experiments}
\vspace{-0.1cm}
In this section, we first show the convergence of inference for BCRF followed by how end-to-end training is performed for a deep network with BCRF. The usefulness of BCRF module is then established through experiments on public datasets. The PQ, RQ, and SQ metrics as defined in \cite{panoptickirillov2017} are used for all experiments. 

\vspace{-0.2cm}
\subsection{Convergence of Inference}
\begin{SCfigure}[][t!]
  \includegraphics[width=0.5\textwidth]{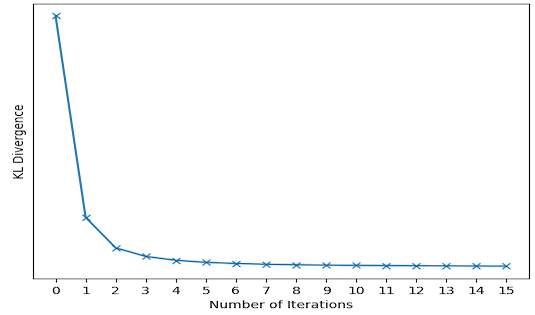}
  \vspace{-0.4cm}
  \caption{\small{{\bf Convergence of BCRF Inference} The KL divergence is plotted against the number of iterations. We pick 20 random images from the Pascal VOC validation set and average the KL divergence for each iteration across these images. It can be seen that the KL divergence measure, and therefore the inference algorithm, converges within a few iterations.}}
  \label{fig:kl_div}
\end{SCfigure}

It is difficult to provide a theoretical convergence guarantee for mean field algorithms with parallel updates~\cite{Higherorder_mf, Koller_book}. We therefore provide empirical evidence by estimating the KL divergence between the original joint distribution and the factorized distribution (see Eq.~\eqref{eq:q_approx}), at the end of each iteration in Algorithm~\ref{alg:infer}. Note that this KL divergence can be estimated up to a constant using the method described in ~\cite{densecrf_suppl}. Our experimental results are shown in Fig.~\ref{fig:kl_div}. We also note that visual results do not change after about 5 iterations.

\begin{SCtable}[][t!]\centering
	\small
	\begin{tabular}{|c|c|c|c|}
		\hline
		\textbf{Method} & \textbf{PQ} & \textbf{SQ} & \textbf{RQ} \\ \hline
		DeeperLab \cite{panoptic_DeeperLab}      & 67.35       & -           & -           \\ \hline
		Ours (baseline) & 70.50       & 88.65       & 78.83       \\ \hline
		Ours (CRF only) & 67.72       & 87.62       & 76.48       \\ \hline
		Ours (BCRF)     & \textbf{71.76}       & \textbf{89.63}       & \textbf{79.33}       \\ \hline
	\end{tabular}
	\caption{\small{\textbf{Results on Pascal VOC Dataset} \\ Our baseline uses DeepLab-v3 and Mask-RCNN followed by combination using the generic method outlined in \cite{panoptickirillov2017}. CRF only corresponds to setting the BCRF cross-potential terms to zero while BCRF is our complete network.}}
	\label{table:Pascal}
	\vspace{-0.2cm}
\end{SCtable}

\vspace{-0.2cm}
\subsection{BCRF in a Deep Network}
In~\cite{Zhen_ICCV15_CRFRNN}, authors show that, in the semantic segmentation setting, mean field inference of a CRF with Gaussian pairwise potentials can be formulated as a Recurrent Neural Network (RNN). Since our BCRF uses an iterative mean field algorithm of similar nature, it is readily adaptable into this RNN based inference described in~\cite{Zhen_ICCV15_CRFRNN}. This formulation allows automatic optimization of the BCRF parameters described in Section~\ref{sec:infer}, using backpropagation and a gradient descent algorithm. Accordingly, we build a PyTorch implementation of BCRF which is used in our experiments. Further, given a suitable loss function for panoptic segmentation, the differentials with respect to this loss can be passed on to both the semantic branch and the instance branch to optimize their parameters, and subsequently, the feature extractor CNN's parameters, thus jointly training the entire network end-to-end.

\vspace{-0.2cm}
\subsection{Results on Pascal VOC Dataset}
In this experiment we use the architecture shown in Figure~\ref{fig:bcrf_net} with generic instance  and semantic segmentation networks, initialize the BCRF parameters with ones obtained through a coarse grid search, and initialize the compatibility matrices as described in Section~\ref{sec:infer}. During both training and inference we used 5 mean-field iterations for BCRF. At the output, we calculate the loss function as a summation of two components: the usual pixel-wise categorical cross entropy loss for the semantic component~\cite{FCN_2015} and the cross entropy loss with "matched" ground truth~\cite{Anurag17} for the instance component. We used full-image training with batch size 1, SGD with learning rate $0.0007$, momentum $0.99$, and run just 10 epochs to obtain the following results. In Table~\ref{table:Pascal}, we report the summary of the quantitative results for the PASCAL VOC validation dataset. Setting cross-potential terms to zero results in a degradation, which highlights the contribution of the BCRF module in fusing two information sources. Qualitative results are shown in Table~\ref{tbl:pascal_visual}.

\vspace{-0.2cm}
\subsection{Results on the COCO Dataset}
\vspace{-0.1cm}

\begin{table}[t!]
\small
\vspace{0.0cm}
\centering
{\arraybackslash
\begin{tabular}{|c|c|c|c|c|c|c|}
	\hline
	& \multicolumn{2}{c|}{\textbf{PQ}} & \multicolumn{2}{c|}{\textbf{SQ}} & \multicolumn{2}{c|}{\textbf{RQ}}    \\ \hline
	\textbf{Category} & Baseline          & BCRF         & Baseline          & BCRF         & Baseline          & BCRF         \\ \hline
	\textbf{All}      & 41.4              & \textbf{41.7}        & 78.3              & \textbf{79.1}         & 50.8              & \textbf{51.1}  \\
	\textbf{Things}   & 47.4          &\textbf{ 47.9}        & 80.4             & \textbf{82.1}         & 57.3              & \textbf{57.7}  \\
	\textbf{Stuff}    & 32.5           & \textbf{33.2}        & 75.1              & \textbf{77.1}         & 40.9              & \textbf{41.6} \\ \hline
\end{tabular}
}
\vspace{0.4cm}
\caption{\small{{\bf Results on COCO dataset.} Comparison of mAP values against a baseline using \cite{panopticFPN}.}}
\label{tbl:coco_values}
\vspace{-0.0 cm}
\end{table}

\begin{table}[t!]
	\small
	\centering
	\begin{tabular}{l|l|l|l|l|l}
		\hline
		\textbf{Method}        & \textbf{Backbone} & \textbf{Params} & \textbf{PQ} & \textbf{PQ\textsuperscript{st}} & \textbf{PQ\textsuperscript{st}} \\ \hline
		OCFusion \cite{InstanceOcclusion}              & ResNetXt-50       & -               & 41.9        & 49.9            & 29.9            \\ \hline
		Panoptic FPN \cite{panopticFPN}           & ResNet - 50       & -               & 39.0        & 45.9            & 28.7            \\ \hline
		Panoptic-DeepLab  \cite{panopticDeepLab}     & Xception-71       & 46.7M           & 41.2        & 44.9            & 35.7            \\ \hline
		Axial-DeepLab \cite{axialdeeplab}         & Axial-ResNet-L    & 44.9M           & 43.9        & 48.6            & 36.8            \\ \hline
		Panoptic FPN with BCRF & ResNet - 50       & 46.0M             & 41.7        & 47.9            & 33.2            \\ \hline
	\end{tabular}
	\vspace{0.4cm}
	\caption{\small{{\bf Comparison with the state-of-the-art for COCO dataset.} We compare against other similar sized networks. Panoptic FPN with BCRF (last row) is our work.}}
	\label{tbl:coco_values}
    \vspace{-0.3cm}
\end{table}

We experiment on the COCO validation set by adding BCRF on top of Panoptic FPN ~\cite{panopticFPN} and training using default parameters and panoptic loss functions in its Detectron2 ~\cite{detectron2} implementation. The quantitative results are reported in Table~\ref{tbl:coco_values}. 

\vspace{-0.2cm}
\subsection{Results on the Cityscapes Dataset}
\vspace{-0.1cm}
To evaluate the usefulness of BCRF without efforts for thorough end-to-end training, we simply plug in BCRF on an existing pre-trained model, followed by fine-tuning on a small subset of train images. We use a COCO pre-trained Panoptic FPN ~\cite{panopticFPN}, and run two experiments (with and without BCRF) training on 200 randomly selected images from the Cityscapes train split. Quantitative results for the entire validation set obtained from this experiment are reported in Table~\ref{tbl:city_values}.

\begin{table}[hbt!]
\small
\vspace{-0.2cm}
\centering
{\arraybackslash
\begin{tabular}{|c|c|c|c|c|c|c|}
	\hline
	& \multicolumn{2}{c|}{\textbf{PQ}} & \multicolumn{2}{c|}{\textbf{SQ}} & \multicolumn{2}{c|}{\textbf{RQ}}    \\ \hline
	\textbf{Category} & Baseline  & BCRF   & Baseline   & BCRF   & Baseline   & BCRF  \\ \hline
	\textbf{All}       & 49.810  & \textbf{50.299}  & 77.271 & \textbf{77.726} & 62.088 & \textbf{62.412} \\
	\textbf{Things} & 46.247 & \textbf{46.547}  & 77.819 &\textbf{ 78.555} & \textbf{59.205} & 59.002 \\
	\textbf{Stuff}   & 52.402  & \textbf{53.028} & 76.872 & \textbf{77.122} & 64.186 & \textbf{64.892} \\ \hline
\end{tabular}
}
\vspace{0.4cm}
\caption{\small{{\bf Results on Cityscapes dataset.} Panoptic segmentation results on the validation set.}}
\label{tbl:city_values}
\vspace{0.0cm}
\end{table}

\begin{figure}[t]
	\begin{center}
		\includegraphics[width=\linewidth]{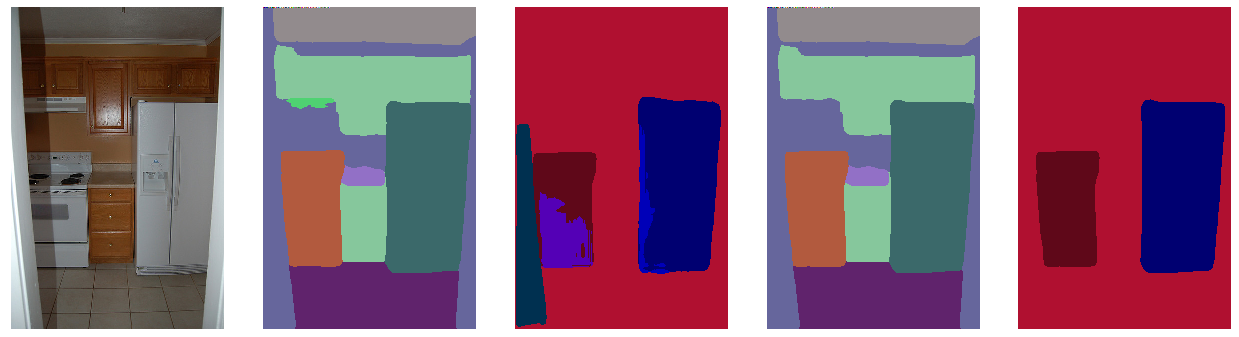}
	\end{center}
	\vspace{-0.1cm}
	\caption{\small{{\bf Visualisation on COCO Dataset.} Columns left to right: original image, semantic output before BCRF, instance output before BCRF, semantic output after BCRF, instance output after BCRF.}}
	\label{fig:vis}
	\vspace{-0.2cm}
\end{figure}

\vspace{-0.8cm}
\subsection{Cross Potentials}
\begin{figure}[hb!]
	\begin{center}
		\includegraphics[width=\linewidth]{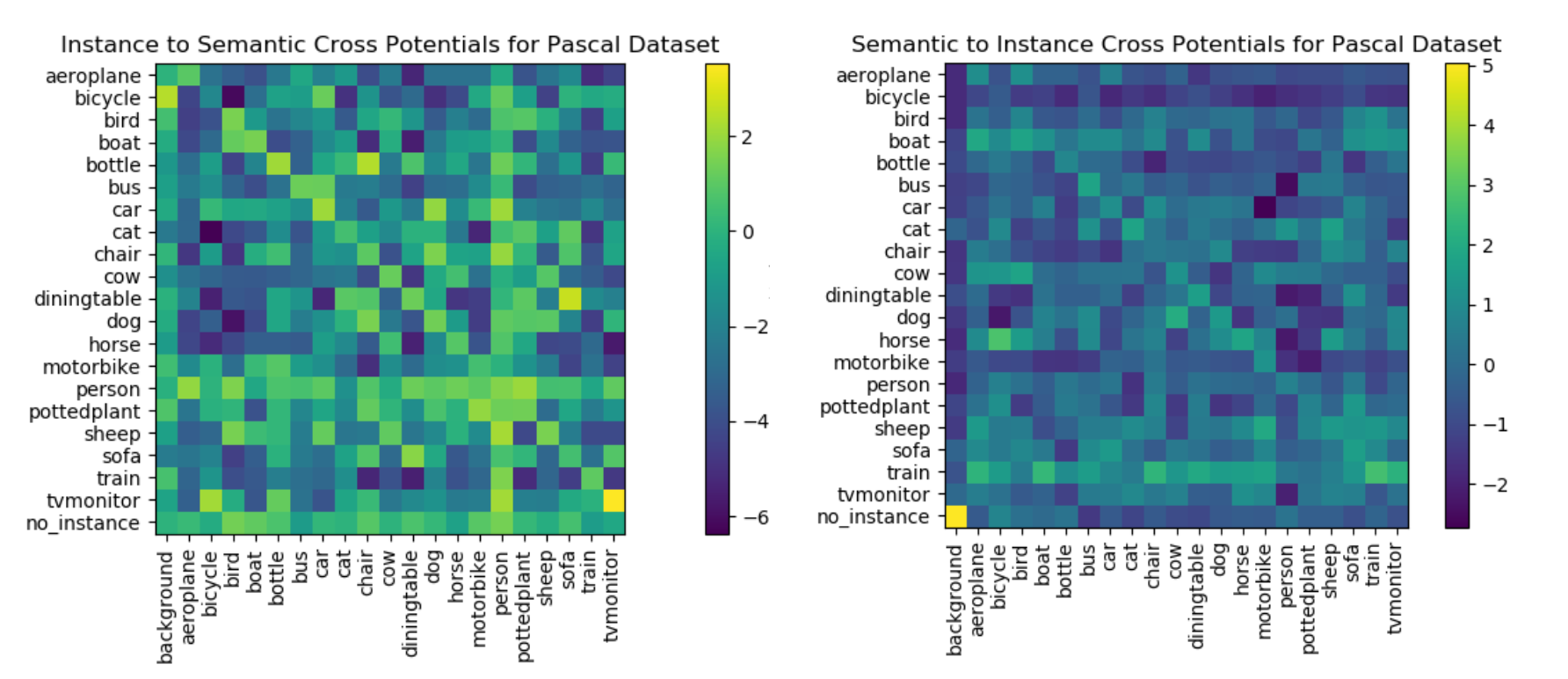}
	\end{center}
	\vspace{-0.2cm}
	\caption{\small{{\bf Heatmap illustrating inter-class dependencies learned by BCRF.}  It shows how important logits belonging to each class in one branch are for predicting each class in the other branch. Instance classes on x-axis and semantic on y-axis. Note that a logarithmic scale has been used for the legend.}}
	\label{fig:potentials}
	\vspace{-0.1cm}
\end{figure}

\vspace{-0.1cm}
Our BCRF module allows the network to learn complex class-aware relationships between the semantic and instance features belonging to each class. While there is room for it to learn a simple logical relationship, the variation of parameters learnt in Figure \ref{fig:potentials} verifies that a complex class-specific mapping has been learned by the network. For example, the high value for dining table to sofa and low value for boat to dining table (Figure \ref{fig:potentials} left) corresponds to the likelihood of finding such objects together in the natural image distribution in Pascal dataset, acting as an attention mechanism on regions of the image. 
\vspace{-0.3cm}
\vspace{-0.5cm}
\section{Conclusion}
\vspace{-0.1cm}
We proposed a probabilistic graphical model based framework for panoptic segmentation. Our BCRF model, containing two different kinds of random variables, is capable of optimally combining the predictions from a semantic segmentation model and an instance segmentation model to obtain a good panoptic segmentation. We use different energy functions in our BCRF to encourage the spatial, appearance, and instance-semantic consistency of the panoptic segmentation. An iterative mean field algorithm is then used to find the panoptic labeling that approximately maximizes the conditional probability of the labeling given the image. We further show that the proposed BCRF framework can be used as an embedded module within a deep neural network to obtain superior results in panoptic segmentation.

\bibliography{egbib}
\end{document}